\documentclass{article}
\usepackage{spconf,amsmath,graphicx}

\usepackage{algorithm}
\usepackage[noend]{algpseudocode}
\usepackage{amssymb}
\usepackage{amsmath}
\usepackage{booktabs}
\usepackage{color}
\usepackage{boldline}
\def\*#1{\mathbf{#1}}
\usepackage[T1]{fontenc}
\usepackage[font=small,labelfont=bf,tableposition=top]{caption}
\DeclareCaptionLabelFormat{andtable}{#1~#2  \&  \tablename~\thetable}
\usepackage[export]{adjustbox}
\usepackage{pifont}% http://ctan.org/pkg/pifont
\usepackage[outline]{contour}% http://ctan.org/pkg/contour
\contourlength{0.01pt}
\contournumber{10}%
\usepackage{wrapfig}
\usepackage{subfig}

\usepackage{arydshln}
\usepackage{url}

\newcommand\blfootnote[1]{%
  \begingroup
  \renewcommand\thefootnote{}\footnote{#1}%
  \addtocounter{footnote}{-1}%
  \endgroup
}

\title{LEARNING GOAL-ORIENTED VISUAL DIALOG VIA TEMPERED POLICY GRADIENT}
\name{Rui Zhao, Volker Tresp}
\address{Ludwig Maximilian University, Oettingenstr. 67, 80538 Munich, Germany \and Siemens AG, Corporate Technology, Otto-Hahn-Ring 6, 81739 Munich, Germany\\}

\begin{document}
%\ninept
%
\maketitle
\begin{abstract}
Learning goal-oriented dialogues by means of deep reinforcement learning has recently become a popular research topic. 
%However, training text-generating agents \textit{efficiently} is still a considerable challenge. 
However, commonly used policy-based dialogue agents often end up focusing on simple utterances and suboptimal policies. To mitigate this problem, we propose a class of novel temperature-based extensions for policy gradient methods, which are referred to as Tempered Policy Gradients (TPGs). 
%These methods encourage exploration with different temperature control strategies. 
%We derive three variations of the TPGs and show their superior performance on a recently published AI-testbed, i.e., the GuessWhat?!\ game. 
On a recent AI-testbed, i.e., the GuessWhat?!\ game, we achieve significant improvements with two innovations. The first one is an extension of the state-of-the-art solutions with Seq2Seq and Memory Network structures that leads to an improvement of 7\%. The second one is the application of our newly developed TPG methods, which improves the performance additionally by around 5\% and, even more importantly, helps produce more convincing utterances.
%TPG can easily be applied to any goal-oriented dialogue systems.
\blfootnote{This paper is an extended version of the IJCAI workshop paper \cite{zhao2018improving}.}
\end{abstract}
\section{Introduction}
\label{sec:intro}

\noindent In recent years, deep learning has shown convincing performance in various areas such as image recognition, %\cite{krizhevsky2012imagenet}, % scene caption \cite{vinyals2015show},
speech recognition, 
%\cite{graves2013speech}
and natural language processing (NLP) \cite{krizhevsky2012imagenet,zhao2017two,graves2013speech,cho2014learning}.
%\cite{cho2014learning}. 
%\cite{lecun2015deep}
Deep neural nets are capable of learning complex dependencies from huge amounts of data and its human generated annotations in a supervised way. In contrast, reinforcement learning agents \cite{sutton1998reinforcement} can learn directly from their interactions with the environment without any supervision in several domains, for instance in the game of GO \cite{silver2016mastering}, as well as many computer games and robotic tasks \cite{mnih2015human,ng2006autonomous, peters2008reinforcement, levine2016end, zhao2018energy}. In this paper we are concerned with the application of both approaches to goal-oriented dialogue systems 
\cite{bordes2016learning,guesswhat_game,visdial_rl,end_to_end_gw,lipton2017bbq,li2016deep,williams2016end,visdial_rl,lewis2017deal,dhingra2016end,zhao2018efficient}, a problem that has recently caught the attention of machine learning researchers. De Vries et al.\  \cite{guesswhat_game} have proposed as AI-testbed a visual grounded object guessing game called GuessWhat?!.  Das et al.\ \cite{visdial_rl} formulated a visual dialogue system which is about two chatbots asking and answering questions to identify a specific image within a group of images.
%Important prior works are  \cite{guesswhat_game}, which have proposed as AI-testbed a visual grounded object guessing game called GuessWhat?! and  \cite{visdial_rl}, which have formulated a visual dialogue system which is about two chatbots asking and answering questions to identify a specific image within a group of images.
%{Here, used new citation format "Authors (year)", so that we have a subject (authors) in a sentence.}
More practically, dialogue agents have been applied to negotiate a deal \cite{lewis2017deal} and access certain information from knowledge bases \cite{dhingra2016end}.
The essential idea in these systems is to train different dialogue agents to accomplish the tasks. In those papers, the agents have been trained with policy gradients, i.e.\ REINFORCE \cite{williams1992simple}.
%either completely autonomously, or supported by human interaction.  \textsc{Comment: what does "multiple" mean here? Do you mean "different"?}{yes}

%%\textsc{Comment: paragraph here}
%One can envision a number of applications of dialogue agents to improve the daily lives of individuals. However, the current state-of-the-art of the goal-oriented dialogue systems is not yet sufficient to warrant commercial applications. We propose that one problem is that the plain policy gradient method, i.e., the REINFORCE algorithm \cite{williams1992simple}, which is used in all current works \cite{end_to_end_gw,visdial_rl,lewis2017deal,dhingra2016end}, is suboptimal for dialogue systems.
%%replaced the word "inadequate" with "flawed"

%The policy gradient approach uses Monte Carlo methods to sample the feedback from actions for model updating, emphasizing the actions that lead to success and penalizing other actions.
%{Consider that, in the context of text generation, the actions are the words selected at each time-step. Experiments have shown that policy-based dialogue agents normally end up producing simple sentences and suboptimal policies \cite{end_to_end_gw}, due to insufficient explorations in the training phase. To mitigate this drawback, we propose a class of methods called Tempered Policy Gradient (TPG), which adjusts the temperatures during sampling in order to find a better balance between exploration and exploitation.

In order to improve the exploration quality of policy gradients, we present three instances of temperature-based methods. The first one is a single-temperature approach which is very easy to apply. The second one is a parallel approach with multiple temperature policies running concurrently. This second approach is more demanding on computational resources,  but results in more stable solutions. The third one is a temperature policy approach that dynamically adjusts the temperature for each action at each time-step, based on action frequencies. This dynamic method is more sophisticated and proves more efficient in the experiments. In the experiments, all these methods demonstrate better exploration strategies in comparison to the plain policy gradient.

We demonstrate our approaches using a real-world dataset called GuessWhat?!. The GuessWhat?!\ game \cite{guesswhat_game} is a visual object discovery game between two players, the Oracle and the Questioner. The Questioner tries to identify an object by asking the Oracle questions. The original works \cite{guesswhat_game,end_to_end_gw} first proposed supervised learning to simulate and optimize the game. Strub et al.\ \cite{end_to_end_gw} showed that the performance could be improved by applying plain policy gradient reinforcement learning, which maximizes the game success rate, as a second processing step.
Building on these previous works, we propose  two  network architecture extensions. We utilize a Seq2Seq model \cite{sutskever2014sequence} to process the image along with the historical dialogues for question generation. For the guessing task, we develop a Memory Network \cite{sukhbaatar2015end} with Attention Mechanism \cite{bahdanau2014neural} to process the generated question-answer pairs. We first train these two models using the plain policy gradient and use them as our baselines.
Subsequently,  we train the models with our new TPG methods and
compare the performances with the baselines. We show that the TPG method is compatible with state-of-the-art architectures such as Seq2Seq and Memory Networks and contributes orthogonally to these advanced neural architectures. To the best of our knowledge, the presented work is the first to propose temperature-based policy gradient methods to leverage exploration and exploitation in the field of goal-oriented dialogue systems. We demonstrate the superior performance of our TPG methods by applying it to the GuessWhat?!\ game. 

\section{Preliminaries}
\label{sec:PG}

In our notation, we use $\*{x}$ to denote the input to a policy network $\pi$, and $x_{i}$ to denote the $i$-th element of the input vector. Similarly, $\*{w}$ denotes the weight vector of $\pi$, and $w_{i}$ denotes the $i$-th element of the weight vector of that $\pi$. The output $y$ is a multinoulli random variable with $N$ states that follows a probability mass function,
\begin{math}
f(y = n \mid \pi (  \*{x} \mid \*{w} ) ),
\end{math}
where $\Sigma_{n=1}^N f(y = n \mid \pi ( \*{x} \mid \*{w} ) ) =1$ and $f(\cdot) \ge 0$. In a nutshell, a policy network parametrizes a probabilistic unit that produces the sampled output, mathematically, $y \sim f(\pi ( \*{x} \mid \*{w} ))$.

%At this point, we have defined the policy neural net and now discuss performance measures commonly used for optimizations. 
Typically, the expected value of the accumulated reward, i.e.\ return, conditioned on the policy network parameters $E(r\mid\*{w})$ is used.
%\textsc{Comment: this expectation $E(r\mid\*{w})$ is very simple; doesn't it look more complex in the literature .... }{used the notation style from (Williams 1992), there are notions involved value function, which looks more complex, but we didn't talk about value funcitons in this paper, only policies. Now, used the term "policy neural nets".}
Here,  $E$ denotes the expectation operator, $r$ the accumulated reward signal, and $\*{w}$ the network weight vector. The objective of reinforcement learning is to update the weights in a way that maximizes the expected return at each trial. In particular, the REINFORCE updating rule is:
\begin{math}\label{eq:DeltaW_pg}
\Delta w_{i} = \alpha_{i} (r-b_{i})e_{i}
\end{math},
where $\Delta w_{i}$ denotes the weight adjustment of weight $w_{i}$, $\alpha_{i}$ is a nonnegative learning rate factor, and $b_{i}$ is a reinforcement baseline. The $e_{i}$ is the \textit{characteristic eligibility} of $w_{i}$, defined as $e_{i} = (\partial f / \partial w_{i}) / f = \partial \mathrm{ln} f / \partial w_{i}$.
%which is the gradient of the probability of taking the action actually taken, divided by the probability of taking that action.
%The name \textit{REINFORCE} is an abbreviation for the updating rule, "\textit{RE}ward \textit{I}ncrement = \textit{N}onnegative \textit{F}actor $\times$ \textit{O}ffset \textit{R}einforcement $\times$ \textit{C}haracteristic \textit{E}ligibility".
Williams \cite{williams1992simple} has proved that the updating quantity, $(r-b_{i})\partial \mathrm{ln}f / \partial w_{i}$, represents an unbiased estimate of $\partial E (r \mid \*{w} ) / \partial w_{i} $.
%the partial derivative of the expected reward function with respect to each weight.

\section{Tempered Policy Gradient}
\label{sec:TPG}
%we have discussed \textsc{in he introduction (?)}{yes, in the intro},

In order to improve the exploration quality of REINFORCE in the task of optimizing policy-based dialogue agents, we attempt to find the optimal compromise between exploration and exploitation. %\textsc{(Comment: Do we have a reference? How know know this? )}{we find that from practices in the work of \cite{end_to_end_gw}}
In TPGs we introduce a parameter $\tau$, the sampling temperature of the  probabilistic output unit, which allows us to explicitly control the strengths of the exploration.
%The functionality of $\tau$ is similar to the $\varepsilon$ in the $\varepsilon$-greedy policy of the Q-learning \cite{sutton1998reinforcement}, in the sense that both parameters control the trade-off between exploration and exploitation.

\subsection{Exploration and Exploitation}
\label{sec:explore}
The trade-off between exploration and exploitation is one of the great challenges in reinforcement learning \cite{sutton1998reinforcement}. 
To obtain a high reward, an agent must exploit the actions that have already proved effective in getting more rewards. However, to discover such actions,
the agent must try  actions, which appear suboptimal, to explore the action space. In a stochastic task like text generation, each action, i.e. a word, must be tried many times to find out whether it is a reliable choice or not. 
The exploration-exploitation dilemma has been intensively studied over many decades \cite{carmel1999exploration,nachum2016improving,liu2017stein}. Finding the balance between exploration and exploitation is considered crucial for the success of reinforcement learning \cite{thrun1992efficient}.

%\subsection{Temperature in Reinforcement Learning}
%The idea of using temperatures in reinforcement learning is not entirely new \cite{sutton1998reinforcement}. For example, Simulated Annealing \cite{kirkpatrick1983optimization} starts with a high temperature (in our case, favoring exploration) and reduces temperature by a specified schedule to eventually achieve pure exploitation. In practice, Simulated Annealing can be painfully slow. Another idea is to add an entropy regularization term in the loss function to encourage exploration \cite{williams1991function}. Unfortunately, none of the existing temperature-based approaches to policy-based agents have proven to be sufficiently effective to be adapted by the deep learning community \cite{goodfellow2016deep}, which was the motivation for our new TPG methods.
%%\textsc{Comment:  this might be a critical; would be great to have a reference. }
%TPG has three variants. We first use a single-temperature method as a basic temperature approach and show that the default $\tau = 1$ might be suboptimal. The second approach uses multiple temperatures in parallel for stabilizing exploration, and the third one uses a frequency-based heuristic to dynamically adjust the temperature. %during exploration.
%%\textsc{Comment: we need to be consistent: either two or three variants?}{three}

\subsection{Temperature Sampling}
\label{sec:temperatureSampling}
In text generation, it is well-known that the simple trick of temperature adjustment is sufficient to shift the language model to be more conservative or more diversified \cite{karpathy2015deep}. In order to control the trade-off between exploration and exploitation, we borrow the strength of the temperature parameter $\tau \ge 0$ to control the sampling. The output probability of each word is transformed by a temperature function as:
\begin{equation*}\label{eq:temFun}
f^{\tau}(y = n \mid \pi ( \*{x} \mid \*{w} ) ) = \frac{f(y = n \mid \pi ( \*{x} \mid \*{w} ) )^{\frac{1}{\tau}}}{\Sigma_{m=1}^N f(y = m \mid \pi ( \*{x} \mid \*{w} ) )^{\frac{1}{\tau}}}   .
\end{equation*}
We use notation $f^{\tau}$ to denote a probability mass function $f$ that is transferred by a temperature function with temperature $\tau$.  
When the temperature is high, $\tau > 1$, the distribution becomes more uniform; when the temperature is low, $\tau < 1$, the distribution appears more spiky. 
\subsection{Tempered Policy Gradient Methods}
\label{sec:TPGs}
%The previous subsections provided an understanding of the temperature parameter and the exploration-exploitation trade-off.
Here, we introduce three instances of TPGs in the domain of goal-oriented dialogues, including single, parallel, and dynamic tempered policy gradient methods.
%These methods take advantage of using different temperature parameters to leverage exploration and exploitation.

\textbf{Single-TPG:}
%\textsc{(Comment: I agree that the eploration/exploitation strategy should have a temperature parameter. I am not sure that this temperature should really  affect the gradients and the updates ... I would assume the policy ia adjusted as before?  )}{temperature parameter does not affect the gradient, the sampled actions do. It is now clarified in single-TPG section.}
The Single-TPG method simply uses a global temperature $\tau_{global}$ during the whole training process, i.e., we use $\tau_{global} > 1$ to encourage exploration. The forward pass is represented mathematically as:
\begin{math}
y^{\tau_{global}} \sim f^{\tau_{global}} ( \pi ( \*{x} \mid \*{w} ) )
\end{math},
where $\pi ( \*{x} \mid \*{w} )$ represents a policy neural network that parametrizes a distribution $f^{\tau_{global}}$ over the vocabulary, and $y^{\tau_{global}}$ means the word sampled from this tempered word distribution. After sampling, the weight of the neural net is updated using,
\begin{equation*}\label{eq:DeltaW_single}
\Delta w_{i} = \alpha_{i} (r-b_{i})\partial \mathrm{ln} f(y^{\tau_{global}} \mid \pi ( \*{x} \mid \*{w} ) ) / \partial w_{i}.
\end{equation*}
Noteworthy is that the actual gradient, $\partial \mathrm{ln} f(y^{\tau_{global}} \mid \pi ( \*{x} \mid \*{w} ) ) / \partial w_{i}$, depends on the sampled word, $y^{\tau_{global}}$, however, does not depend directly on the temperature, $\tau$. 
%We prefer to find the words that lead to a reward, so that the model can learn quickly from these actions, otherwise, the neural network only learns to avoid current failure actions.
%, which would slow down the learning process.
%\textsc{(Comment: I do not completely understand the last sentence)}{agents learn fast with reward, learn slow with penalization}
With Single-TPG and $\tau > 1$, the entire vocabulary of a dialogue agent is explored more efficiently than by REINFORCE, because nonpreferred words have a higher probability of being explored. 
%This Single-TPG method is very easy to use and could yield a performance improvement after training because the goal-oriented dialogue optimization could benefit from increased exploration. 
%The temperature is initialized  with $\tau = 1$, then fine-tuned based on the learning curve on the validation sets, and subsequently  left fixed..
%\textsc{(Comment: using cross validation or how?)}{yes using cross validation}
%\textsc{ Comment: I would remove the next sentence: This selection process indicates that in the worse case TPG performs as well as REINFORCE does.}{removed}
%\textsc{Comment: we need to write something about how $\tau$ is selected / optimized}{added}

\textbf{Parallel-TPG:}
A more advanced version of Single-TPG is the Parallel-TPG that deploys several Single-TPGs concurrently with different temperatures, $\tau_1, ..., \tau_n$, and updates the weights based on all generated samples. During the forward pass, multiple copies of the neural nets parameterize multiple word distributions. The words are sampled in parallel at various temperatures, mathematically,
\begin{math}
y^{\tau_1}, ..., y^{\tau_n} \sim f^{\tau_{1}, ..., \tau_{n} } (\pi ( \*{x} \mid \*{w} ) )
\end{math}.
After sampling, in the backward pass the weights are updated with the sum of gradients. The formula is given by
\begin{equation*}\label{eq:DeltaW_async}
\Delta w_{i} = \Sigma_{k} \alpha_{i} (r-b_{i})\partial \mathrm{ln} f(y^{\tau_{k}} \mid \pi ( \*{x} \mid \*{w} ) ) / \partial w_{i},
\end{equation*}
where $k\in \{1, ..., n\}$. The combinational use of higher and lower temperatures ensures both exploration and exploitation at the same time. The sum over weight updates of parallel policies gives a more accurate Monte Carlo estimate of $\partial E (r \mid \*{w} ) / \partial w_{i} $, due to the nature of Monte Carlo methods \cite{robert2004monte}. Thus, compared to Single-TPG, we would argue that Parallel-TPG is more robust and stable, although Parallel-TPG needs more computational power.
%\textsc{Comment: how do we know that? From the experiments?}{due to the nature of Monte Carlo methods, the more you sampled, the more accurate it is.}
However, these computations can easily be distributed in a parallel fashion using state-of-the-art graphics processing units.

\textbf{Dynamic-TPG:}
As a third variant, we introduce the Dynamic-TPG, which is the most sophisticated approach in the current TPG family. The essential idea is that we use a heuristic function $h$ to assign the temperature $\tau$ to the word distribution at each time step, $t$. The temperature is bounded in a predefined range $[\tau_{min}, \tau_{max}]$. The heuristic function we used here is based upon the term frequency inverse document frequency, $\textit{tf-idf}$ \cite{leskovec2014mining}. In the context of goal-oriented dialogues, we use the counted number of each word as term frequency \textit{tf} and the total number of generated dialogues during training as document frequency \textit{df}. We use the word that has the highest probability to be sampled at current time-step, $y_{t}^{*}$, as the input to the heuristic function $h$. Here,  $y_{t}^{*}$ is the maximizer of the probability mass function $f$. Mathematically, it is defined as
\begin{math}
y_{t}^{*} = \mathrm{argmax} ( f ( \pi ( \*{x} \mid \*{w} ) ) )
\end{math}.
We propose that $\textit{tf-idf}(y_{t}^{*})$ approximates the concentration level of the distribution, which means that if the same word is always sampled from a distribution, then the distribution is very concentrated. Too much concentration prevents the model from exploration, so that a higher temperature is needed.
In order to achieve this effect, the heuristic function is defined as
\begin{equation*}\label{eq:tau_heuristic}
\begin{split}
\tau_{t}^h &= h(\textit{tf-idf}(y_{t}^{*})) \\
	   &= \tau_{min} + (\tau_{max} - \tau_{min}) \frac{\textit{tf-idf}(y_t^*)- \textit{tf-idf}_{min}}{\textit{tf-idf}_{max}- \textit{tf-idf}_{min}}.
\end{split}
\end{equation*}
%\begin{equation*}\label{eq:tau_heuristic}
%\tau_{t}^h = h(\textit{tf-idf}(y_{t}^{*}))
%	   = \tau_{min} + (\tau_{max} - \tau_{min}) \frac{\textit{tf-idf}(y_t^*)- \textit{tf-idf}_{min}}{\textit{tf-idf}_{max}- \textit{tf-idf}_{min}}.
%\end{equation*}
With this heuristic, words that occur very often 
are depressed by applying a higher temperature to those words, making them less likely to be selected in the near future.
%With this heuristic, the words that occur too often are depressed by applying a higher temperature to that word distribution.
%\textsc{Comment: I am not completely following this but if we have word specific temperatures, don't we need a work index somewhere. $\tau^h$ looks word independent.}{$\tau$ is also time dependent, because each word at each time step. so add time index t }
In the forward pass, a word is sampled using
\begin{math}
y^{\tau_t^h} \sim f^{\tau_t^{h}} ( \pi ( \*{x} \mid \*{w} ) )
\end{math}.
In the backward pass, the weights are updated correspondingly:
\begin{equation*}\label{eq:DeltaW_dync}
\Delta w_{i} = \alpha_{i} (r-b_{i})\partial \mathrm{ln} f(y^{\tau_t^{h}} \mid \pi ( \*{x} \mid \*{w} ) ) / \partial w_{i},
\end{equation*}
where $\tau_t^h$ is the temperature calculated from the heuristic function. Compared to Parallel-TPG, the advantage of Dynamic-TPG is that it assigns temperature more appropriately, without increasing the computational load.
%\textsc{Comment: why does single-TPG need computational power?}{only Compared to Parallel-TPG}

\section{GuessWhat?!\ Game}
\label{sec:guesswhat}
We evaluate our methods using  a recent testbed for AI, called the GuessWhat?!\ game \cite{guesswhat_game}, available at \url{https://guesswhat.ai}. The dataset consists of 155\,k dialogues, including 822\,k question-answer pairs, each composed of around 5\,k words, about 67\,k images \cite{lin2014microsoft} and 134\,k objects. The game is about visual object discovery trough a multi-round QA among different players.

%GuessWhat?!\ is a two-player guessing game in which both players see a natural picture with a rich number of objects. One player - the Oracle - selects one specific object within the picture. The other player - the Questioner - tries to find out which object the Oracle has selected.
%%The Questioner is allowed to ask the Oracle a series of yes-no questions regarding the objects in the image. Once the Questioner is ready, it takes a guess about the object and locates the specific object in the picture. If it points out the object correctly, then the game is completed successfully; otherwise it counts as failure.
%To play the game, de Vries et al.\ define the roles of the players. The Oracle's task is to give a yes-no answer to the Questioner's questions. De Vries et al.\ divide the Questioner's task into two parts. The first part is the Question-Generator model, in which the task is to ask meaningful questions about the objects. The second part is the Guesser model, in which a guess is taken only once at the end of the dialogue.

Formally, a GuessWhat?!\ game is represented by a tuple $(I, D, O, o^*)$, where $I \in \mathbb{R}^{H \times W}$ denotes an image of height $H$ and width $W$; $D$ represents a dialogue composed of $M$ rounds of question-answer pairs (QAs), $D = (\*{q}_m, a_m)_{m=1}^M$; $O$ stands for  a list of $K$ objects $O = (o_k)_{k=1}^K$; and $o^*$ is the target object. Each question is a sequence of words, $\*{q}_m = \{y_{m,1},......,y_{m,N_m}\}$ with length $N_m$. The words are taken from a defined vocabulary $V$, which consists of the words and  a start token and an end token. Each answer is either yes, no, or not applicable, i.e. $a_m \in \{yes, no, n.a.\}$. For each object $o_k$, there is a corresponding object category $c_k \in  \{1, ......, C \}$ and a pixel-wise segmentation mask $S_k \in \{0, 1\}^{H \times W}$. Finally, we use colon notation ($:$) to select a subset of a sequence, for instance, $(\*{q}, a)_{1:m}$ refers to the first $m$ rounds of QAs in a dialogue.

\subsection{Models and Pretraining}
\label{sec:3models}
Following \cite{end_to_end_gw}, we first train all three models in a supervised fashion.
%The details about the model definition for the Oracle, question-generator, and guesser models are following.

\textbf{Oracle:}
\begin{figure}%[thpb], figure* for take two columns
	\centering
	\includegraphics[width=2.3 in]{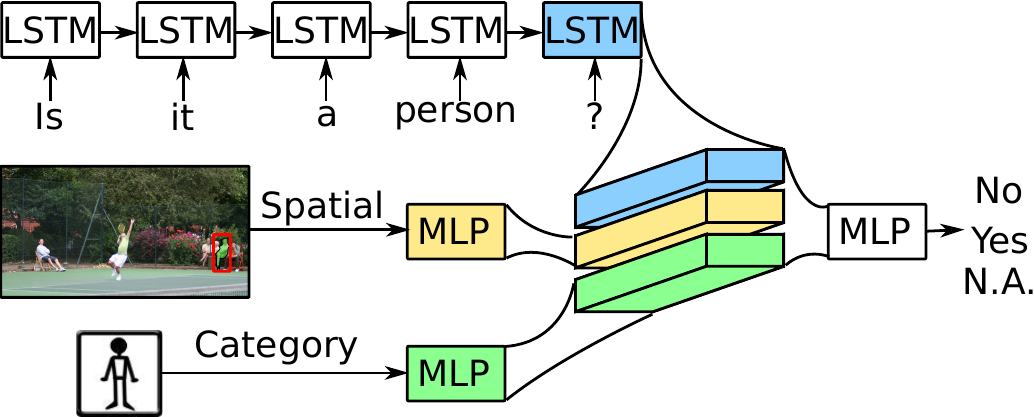}
	%\missingfigure[figwidth=8cm]{Place holder}
	\caption{Oracle model}
	\label{fig:Oracle}
\end{figure}
The task of the Oracle is to answer questions regarding to the target object. We outline here the simple neural network architecture that achieved the best performance in the study of \cite{guesswhat_game}, and which we also used in our experiments. The input information used here is of three modalities, namely the question $\*{q}$, the spatial information $x_{spatial}^*$ and the category $c^*$ of the target object. For encoding the question, de Vries et al.\ first use a lookup table to learn the embedding of words, then use a one layer long-short-term-memory (LSTM) \cite{hochreiter1997long} to encode the whole question. For spatial information, de Vries et al.\ extract an 8-dimensional vector of the location of the bounding box $[x_{min},\ y_{min},\ x_{max},\ y_{max},\ x_{center},\ y_{center},\ w_{box},\ h_{box}]$, where $x$, $y$ denote the coordinates and $w_{box}$, $h_{box}$ denote the width and height of the bounding box, respectively. De Vries et al.\ normalize the image width and height so that the coordinates range from -1 to 1. The origin is at the image center. The category embedding of the object is also learned with a lookup table during training. At the last step, de Vries et al.\ concatenate all three embeddings into one feature vector and fed it into a one hidden layer multilayer perceptron (MLP). The softmax output layer predicts the distribution,
\begin{math}
\mathrm{Oracle} :=  p(a\mid\*{q},\ c^*,\ x_{spatial}^*)
\end{math},
over the three classes, including no, yes, and not applicable.
The model is trained using the negative log-likelihood criterion. The Oracle structure is shown in Fig.~\ref{fig:Oracle}.
% \textsc{Comment: Shouldn't we mention the MLP and describe its architecture: how many layers ... }{mentioned above MLP}

\textbf{Question-Generator:}
\begin{figure}%[thpb], figure* for take two columns
	\centering
	\includegraphics[width=2.8 in]{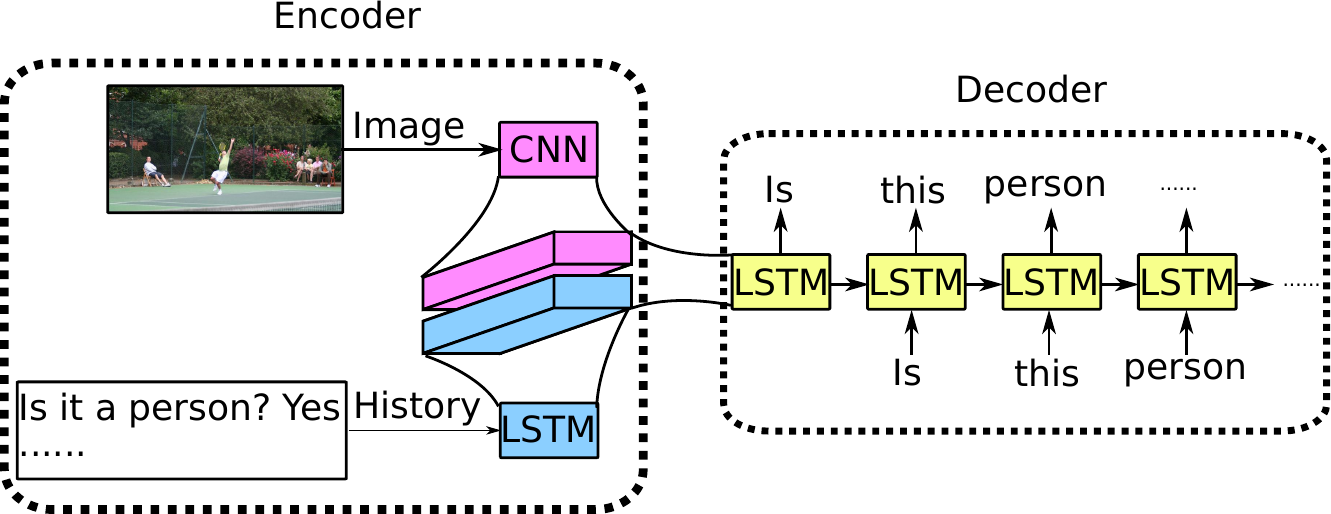}
	%\missingfigure[figwidth=8cm]{Place holder}
	\caption{Question-Generator model}
	\label{fig:qgen}
\end{figure}
%We divide the task of the Questioner into two parts, the question-generator (QGen) and the guesser model. Here we illustrate the former in detail.
The goal of the Question-Generator (QGen) is to ask the Oracle meaningful questions, $\*{q}_{m+1}$, given the whole image, $I$, and the historical question-answer pairs, $(\*{q},a)_{1:m}$. In previous work \cite{end_to_end_gw}, the state transition function was modelled as an LSTM, which was trained using whole dialogues so that the model memorizes the historical QAs. We refer to this as dialogue level training.
%\textsc{ Comment: is all of the following new? You still need to be more clear of what was there before (\cite{end_to_end_gw}) and what is your innovation!}{modified}
We develop a novel QGen architecture using a modified version of the Seq2Seq model \cite{sutskever2014sequence}. The modified Seq2Seq model enables \textit{question level training}, which means that the model is fed with historical QAs, and then learns to produce a new question. %\textsc{(Comment: I don't understand " history and question in pair")}{rephrased the above sentence}
%This process repeats one question by one question instead of one dialogue. \textsc{(Comment: I do not understand the last sentence)}{deleted the last sentence. It means iterate over batches of quesitons not batches of dialogues}
Following \cite{end_to_end_gw}, we first encode the whole image into a fixed-size feature vector using the VGG-net \cite{simonyan2014very}. The features come from the fc-8 layer of the VGG-net. For processing historical QAs, we use a lookup table to learn the word embeddings, then again use an LSTM encoder to encode the history information into a fixed-size latent representation, and concatenate it with the image representation:
\begin{equation*}
\*{s}^{enc}_{m,Nm} = \mathrm{encoder}((\mathrm{LSTM}(\*{q},a)_{1:m}), \mathrm{VGG}( I)).
\end{equation*}
The encoder and decoder are coupled by initializing the decoder state with the last encoder state, mathematically,
\begin{math}
\*{s}^{dec}_{m+1,0} = \*{s}^{enc}_{m,Nm}
\end{math}.
The LSTM decoder generates each word based on the concatenated representation and the previous generated word (note the first word is a start token):
\begin{equation*}
y_{m+1,n} = \mathrm{decoder}(\mathrm{LSTM}((y_{m+1,n-1}, \*{s}^{dec}_{m+1,n-1})).
\end{equation*}
The decoder shares the same lookup table weights as the encoder. The Seq2Seq model, consisting of the encoder and the decoder, is trained end-to-end to minimize the negative log-likelihood cost. During testing, the decoder gets a start token and the representation from the encoder, and then generates each word at each time step until it encounters a question mark token,
\begin{math}
\mathrm{QGen} : = p(y_{m+1,n}\mid(\*{q},a)_{1:m},\ I)
\end{math}.
The output is a complete question. After several question-answer rounds, the QGen outputs an end-of-dialogue token, and stops asking questions.
%One round of dialogue is finished. \textsc{(Comment: don't understand the last sentence)}
The overall structure of the QGen model is illustrated in Fig.~\ref{fig:qgen}.

\textbf{Guesser:}
\begin{figure}%[thpb], figure* for take two columns
	\centering
	\includegraphics[width=2.9 in]{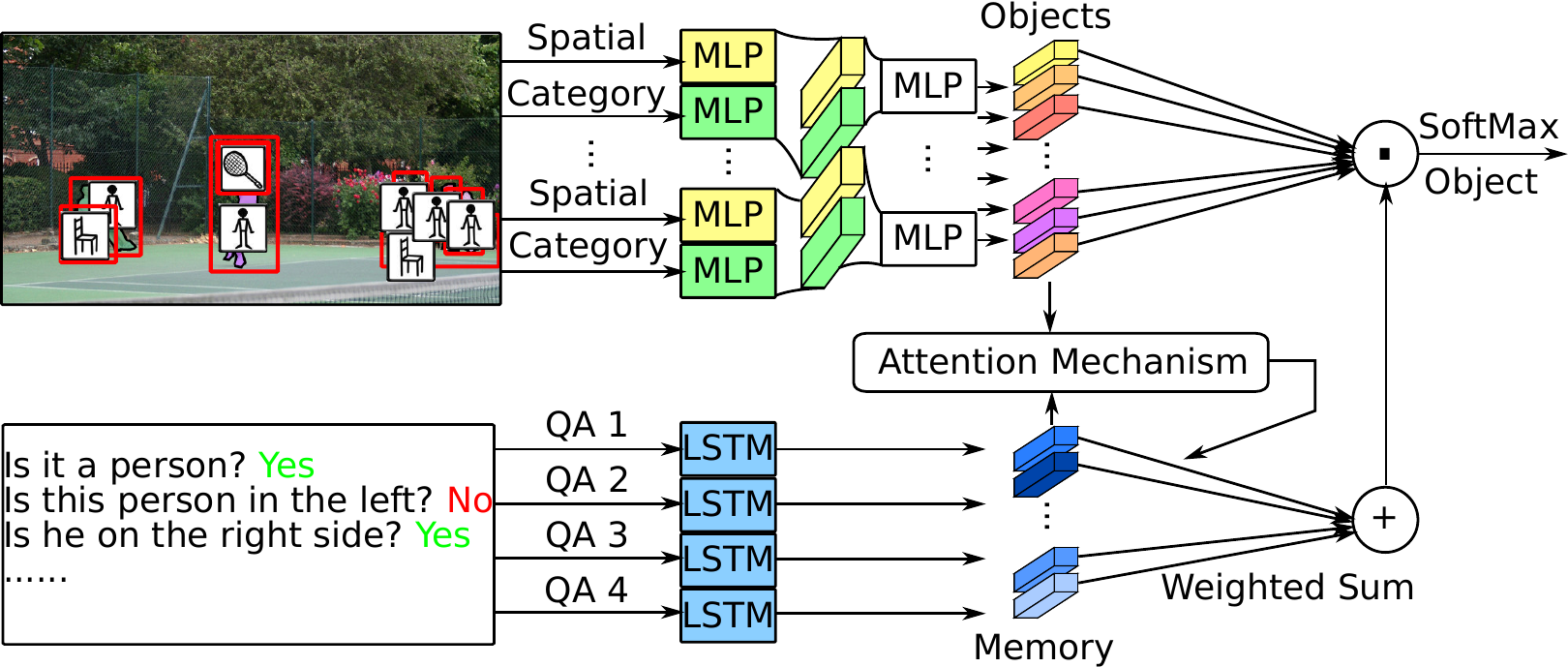}
	%\missingfigure[figwidth=8cm]{Place holder}
	\caption{Guesser model}
	\label{fig:guesser}
\end{figure}
%We split the task of the Questioner into two parts. The second part is the guesser model.
The goal of the Guesser model is to find out which object the Oracle model is referring to, given the complete history of the dialogue and a list of objects in the image, $p(o^*\mid(\*{q},a)_{1:M},\ x^{O}_{spatial},\ c^{O})$. The Guesser model has access to the spatial, $x^{O}_{spatial}$, and category information, $c^{O}$,  of the objects in the list. The task of the Guesser model is challenging because it needs to understand the dialogue and to focus on the important content, and then guess the object. To accomplish this task, we decided to integrate the Memory \cite{sukhbaatar2015end} and Attention \cite{bahdanau2014neural} modules into the Guesser architecture used in the  previous work \cite{end_to_end_gw}. First, we use an LSTM header to process the varying lengths of question-answer pairs in parallel into multiple fixed-size vectors. Here, each QA-pair has been encoded into some facts,
\begin{math}
\mathrm{Fact}_m = \mathrm{LSTM}((\*{q},a)_{m})	
\end{math},
and stored into a memory base.
Later, we use the sum of the spatial and category embeddings of all objects as a key,
\begin{math}
\mathrm{Key}_1 = \mathrm{MLP}(x^{O}_{spatial},\ c^{O})
\end{math},
to query the memory and calculate an attention mask,
\begin{math}
\mathrm{Attention}_1(\mathrm{Fact}_m) = \mathrm{Fact}_m \odot  \mathrm{key_1}
\end{math},
over each fact.
Next, we use the sum of attended facts and the first key to calculate the second key. Further, we use the second key to query the memory base again to have a more accurate attention. These are the so called ``two-hops" of attention in the literature \cite{sukhbaatar2015end}.  Finally, we compare the attended facts with each object embedding in the list using a dot product. The most similar object to these facts is the prediction, \begin{math}
\mathrm{Guesser} := p(o^*\mid(\*{q},a)_{1:M},\ x^{O}_{spatial},\ c^{O})
\end{math}.
The intention of using the attention module here is to find out the most relevant descriptions or facts concerning the candidate objects. We train the whole Guesser network end-to-end using the negative log-likelihood criterion. A more graphical description of the Guesser model is shown in Fig.~\ref{fig:guesser}.

\subsection{Reinforcement Learning}
\label{sec:RF}
%Now, we have all three models defined and pre-trained in a supervised fashion. We can simulate the whole game and assign rewards to the models.
Now, we post-train the QGen and the Guesser model with reinforcement learning. We keep the Oracle model fixed.
%For a full game episode, the QGen first asks a question about the object in the image. The Oracle model gives a yes-no answer to the question. After several rounds of question-answering, the QGen decides to stop asking. Finally, the guesser model reads the whole dialogue and predicts the object. When the guesser model finds the correct one, then it gets one as a reward, $r = 1$, when it takes the wrong guess, it gets zero as a reward, $r = 0$.
In each game episode, when the models find the correct object, $r = 1$, otherwise, $r = 0$.

Next, we can assign credits for each action of the QGen and the Guesser models. In the case of the QGen model, we spread the reward uniformly over the sequence of actions in the episode. The baseline function, $b$, used here is the running average of the game success rate. Consider that the Guesser model has only one action in each episode, i.e., taking the guess. If the Guesser finds the correct object, then it gets an immediate reward and the Guesser's parameters are updated using the REINFORCE rule without baseline.
The  QGen is trained  using the following four methods.

\textbf{REINFORCE:}
The baseline method used here is REINFORCE \cite{williams1992simple}. During training, in the forward pass the words are sampled with $\tau = 1$,
\begin{math}
y_{m+1, n} \sim f ( \mathrm{QGen}(\*{x} \mid \*{w}))
\end{math}.
In the backward pass, the weights are updated using REINFORCE,
$
\*{w} = \*{w} + \alpha (r - b) \nabla_{\*{w}} \mathrm{ln} f(y_{m+1, n} \mid \mathrm{QGen}(\*{x} \mid \*{w}) ).
$

\textbf{Single-TPG:}
We use temperature $\tau_{global} = 1.5$ during training to encourage exploration, mathematically,
$
y^{\tau_{global}}_{m+1, n} \sim f^{\tau_{global}} ( \mathrm{QGen}(\*{x} \mid \*{w}))
$.
In the backward pass, the weights are updated using
$
\*{w} = \*{w} + \alpha (r - b) \nabla_{\*{w}} \mathrm{ln} f(y^{\tau_{global}}_{m+1, n} \mid \mathrm{QGen}(\*{x} \mid \*{w}) ).
$

\textbf{Parallel-TPG:}
For Parallel-TPG, we use two temperatures $\tau_1 = 1.0$ and $\tau_2 = 1.5$ to encourage the exploration. The words are sampled in the forward pass using
$
y^{\tau_1}_{m+1, n},\ y^{\tau_2}_{m+1, n} \sim f^{\tau_1, \tau_2} ( \mathrm{QGen}(\*{x} \mid \*{w}))
$.
In the backward pass, the weights are updated using
$
\*{w} = \*{w} + \Sigma_{k=1}^2 \alpha (r - b) \nabla_{\*{w}} \mathrm{ln} f(y^{\tau_{k}}_{m+1, n} \mid \mathrm{QGen}(\*{x} \mid \*{w}) ) .
$

\textbf{Dynamic-TPG:}
The last method we evaluated  is Dynamic-TPG. We use a heuristic function to calculate the temperature for each word at each time step:
$
\tau_{m+1,n}^h
	   = \tau_{min} + (\tau_{max} - \tau_{min}) \frac{\textit{tf-idf}(y_{m+1,n}^*)- \textit{tf-idf}_{min}}{\textit{tf-idf}_{max}- \textit{tf-idf}_{min}},
$
where we set $\tau_{min} = 0.5$, $\tau_{max} = 1.5$, and set $\textit{tf-idf}_{min} = 0$, $\textit{tf-idf}_{max} = 8$. After the calculation of $\tau_{m+1,n}^h$, we substitute the value into the formula at each time step and sample the next word using
$
y^{\tau_{m+1,n}^h}_{m+1, n} \sim f^{\tau_{m+1,n}^h} ( \mathrm{QGen}(\*{x} \mid \*{w})).
$
In the backward pass, the weights are updated using
$
\*{w} = \*{w} + \alpha (r - b) \nabla_{\*{w}} \mathrm{ln} f(y^{\tau_{m+1,n}^h}_{m+1, n} \mid \mathrm{QGen}(\*{x} \mid \*{w}) ) .
$
For all four methods, we use greedy search in evaluation. %We can also use methods such as sampling, greedy search, and beam-search to enhance the model \cite{jia2015guiding}.
%\textbf{Full Training Process}
%W summarise the full training process using Tempered Policy Gradient (TPG) in Algo.~\ref{algo:full_train}.
%\makeatletter
%%\def\BState{\State\hskip-\ALG@thistlm}
%\makeatother
%\begin{algorithm}
%\caption{Training of QGen with TPG}\label{algo:full_train}
%\begin{algorithmic}[1]
%\Require{Pretrained Oracle, QGen and Guesser}
%\Require{Batch size K}
%\For{Each update}
%\State \# Generate trajectories $T$
%	\For{$k =1\  \text{to}\  K$}
%	\State Pick Image $I_{k}$ and the target object $o_{k}^{*} \in O_{k}$
%	\State \# Generate question-answer pairs $(\*{q},a)_{1:m}^{k}$
%	\State Update QGen temperatures with Eq.()
%		\For{$m = 1 \text{ to } M_{max}$}
%		\State $q_m^k = \mathrm{QGen}((\*{q},a)^{k}_{1:m-1}, I_k)$
%		\State $a_m^k = \mathrm{Oracle}(\*{q}, o_k^*, I_k) $
%			\If {$ <\textit{END}> \in \*{q}_i^k$}
%			\State delete $(q,a)_n^k$ and break;
%			\EndIf
%		\EndFor
%		\State $p(o_k\mid\cdot) = \mathrm{Guesser}((\*{q}, a)_{1:m}^k, I_k, O_k)$
%		\State $r_k =
%			\begin{cases}
%			1		& \text{if } \mathrm{argmax_{o_{k}}} p(o_k\mid\cdot) = o_k^*\\
%			0   	& \text{otherwise }
%			\end{cases}
%		$
%	\EndFor
%	\State Define $T = ( (\*{q},a)_{1:m_k}^{k}, I_{k}, r_k )_{1:K} $
%	\State Update baseline with accuracy running average
%	\State Update $\textit{tf-idf}$ for each word
%	\State Evaluate $\nabla J(\theta_h)$ with Eq. with $T$
%	\State SGD update of Guesser parameters $\theta$ using $\nabla J(\theta_h)$
%	\State Evaluate $\nabla L (\phi)$ with Eq. with $T$
%	\State SGD update of QGen parameters using $\nabla L (\phi)$
%\EndFor
%\end{algorithmic}
%\end{algorithm}

\section{Experiment}
%\begin{table}%[thpb] * takes two-columns
%\centering
%\begin{tabular}{
%p{3.8cm}  					p{.9cm}  		p{.9cm}  	p{.9cm}} 	\hlineB{3}
%Method 					& 	Oracle 		&	QGen\textsuperscript{3}		& 	Guesser	\\ \hline
%\cite{guesswhat_game} 	&	78.5\%		& 46.8\% 	& 61.3\% 	\\
%\cite{end_to_end_gw}\textsuperscript{1}		&	78.5\%		& 44.6\% 	& 63.8\% 	\\
%\cite{guesswhat_github}\textsuperscript{2}  &	78.9\%		& 44.6\% 	& 64.2\% 	\\
%Our method				&	78.72\%		& \textbf{48.77\%} 	& 67.12\% 	\\ \hlineB{3}
%\end{tabular}
%\raggedright \small \ \textsuperscript{2} is a fine-tuned version of \ \textsuperscript{1}. \ \textsuperscript{3} our method uses greedy search only, others also use beam-search to improve the performance.
%\caption{Performance comparison after supervised learning}
%\label{tab:supervised}
%\end{table}
We first train all the networks in a supervised fashion, and then optimize the QGen and the Guesser model using reinforcement learning. 
Our implementation \footnote{https://github.com/ruizhaogit/GuessWhat-TemperedPolicyGradient} uses Torch \cite{torch}. 
%The source code is available at this link \footnote{https://github.com/ruizhaogit/GuessWhat-TemperedPolicyGradient}.
%, which uses Torch7 \cite{torch}. 
\subsection{Pretraining}
We train all three models using 0.5 dropout \cite{srivastava2014dropout} during training, using the ADAM optimizer \cite{kingma2014adam}. We use a learning rate of 0.0001 for the Oracle model and the Guesser model, and a learning rate of 0.001 for QGen. All the models are trained with at most 30 epochs and early stopped within five epochs without improvement on the validation set. We report the performance on the test set which consists of images not used in training.
% After supervised training, the Oracle model obtains 78.72\% accuracy, and the guesser model reports 67.12\% on the test set. The QGen model obtains 48.77\% game success rate in cooperating with the Oracle and the guesser models. During testing, we only use greedy search to sample the words.
We report the game success rate as the performance metric for all three models, which equals to the number of succeeded games divided by the total number of all games. %\textsc{Comment: what is that?}{you mean what is success rate?}.
%A comparison of our methods and previous methods is shown in Tab.~\ref{tab:results}.
Compared to previous works \cite{guesswhat_game,end_to_end_gw,guesswhat_github}, 
%our reimplementation of the Oracle model obtains a similar performance. Our new Guesser model performs 3\% better than the best model from literature. Our QGen model improves the accuracy by 4\% compared to \cite{end_to_end_gw}, which we attribute to the better Guesser model. 
after supervised training, our models obtain a game success rate of 48.77\%, that is 4\% higher than state-of-the-art methods \cite{guesswhat_github}, which has 44.6\% accuracy. 
%\textsc{ Comment: should'nt this sentence come earlier?}

%\textsc{Comment: where do I see the 15\%?}{it is in the RF part}

\subsection{Reinforcement Learning}
\begin{table}%[thpb] * takes two-columns
\centering
\begin{tabular}{ p{0.5cm}    p{5.1cm} 	p{1.2cm} } \hlineB{3}
\# 	& Method 																& Accuracy 					\\ \hline
1 	& Strub et al., 2017 \cite{end_to_end_gw}									& 52.30\% 					\\
2 	& Strub and de Vries, 2017 \cite{guesswhat_github}						& 60.30\% 					\\
3 	& Our Torch reimplementation of (\#\,2)								& \textbf{62.61}\% 					\\
4 	& (\#\,3) + new QGen (Seq2Seq)												& 63.47\% 					\\
5 	& (\#\,4) + new Guesser (Memory Nets)										& 68.32\%					\\
6 	& (\#\,5) + new Guesser (REINFORCE) 		& \textbf{69.66\%}			\\
7 	& (\#\,6) + Single-TPG																& 69.76\% 					\\
8 	& (\#\,6) + Parallel-TPG															& 73.86\% 					\\
9 	& (\#\,6) + Dynamic-TPG															& \textbf{74.31\%} 			\\ \hlineB{3}
\end{tabular}
\caption{Performance comparison and ablation tests}
\label{tab:results}
\end{table}
%After training all three models in a supervised fashion, we post-train the QGen and the guesser model jointly using reinforcement learning.
We first initialize all models with pre-trained parameters from supervised learning and then post-train the QGen using either REINFORCE or TPG for 80 epochs. We update the parameters using stochastic gradient descent (SGD) with a learning rate of 0.001 and a batch size of 64. In each epoch, we sample each image in the training set once and randomly pick one of the objects as a target.  We track the running average of the game success rate and use it directly as the baseline, $b$, in REINFORCE.
%\textsc{Comment: I do not understand lastsentence}
We limit the maximum number of questions to 8 and the maximum number of words to 12. Simultaneously, we train the Guesser model using REINFORCE without baseline and using SGD with a learning rate of 0.0001. The performance comparison is shown in Tab.~\ref{tab:results}.

%\begin{table}[t]%[thpb] * takes two-columns
%\centering
%\begin{tabular}{
%p{6cm}  								p{1.2cm} 		} \hlineB{3}
%Method 								& Accuracy 		\\ \hline
%REINFORCE							& 73.47\% 		\\
%Single-TPG							& 74.56\% 		\\
%Parallel-TPG							& 75.63\% 		\\
%Dynamic-TPG							& \textbf{75.93\%} 		\\ \hlineB{3}
%\end{tabular}
%\caption{Comparison of the TPGs with REINFORCE}
%\label{tab:tempered}
%\end{table}

\begin{table*}
\centering
\smallskip%\noindent
\begin{tabular}{ c  p{5cm}  p{5cm}  }
\toprule
Image & Policy Gradient & Tempered Policy Gradient \\
\cmidrule(r){1-1}\cmidrule(lr){2-2}\cmidrule(l){3-3}
\raisebox{-\totalheight}{\includegraphics[width=1.8 in]{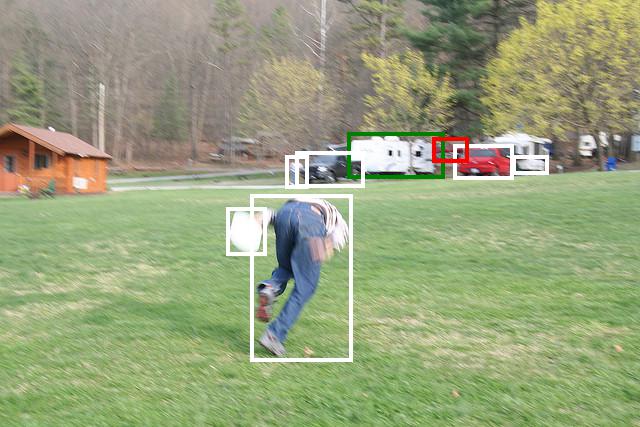}}
&
{\fontsize{9}{10}\selectfont
Is it in left? \hfill{No} \newline
Is it in front? \hfill{No} \newline
Is it in right? \hfill{Yes} \newline
Is it in middle? \hfill{Yes} \newline
Is it person? \hfill{No} \newline
Is it ball? \hfill{No} \newline
Is it bat? \hfill{No} \newline
Is it \textbf{car}? \hfill{Yes} \newline
Status: \hfill{\textcolor{red}{Failure}}
}	
&
{\fontsize{9}{10}\selectfont
Is it a person? \hfill{No} \newline
Is it a \textbf{vehicle}? \hfill{Yes} \newline
Is it a \textbf{truck}? \hfill{Yes} \newline
Is it in front of photo? \hfill{No} \newline
In the left half? \hfill{No} \newline
In the middle of photo? \hfill{Yes} \newline
Is it to the right photo? \hfill{Yes} \newline
Is it in the middle of photo? \hfill{Yes} \newline
Status: \hfill{\textcolor{green}{Success}}
}
\\ \midrule
\raisebox{-\totalheight}{\includegraphics[width=1.8 in]{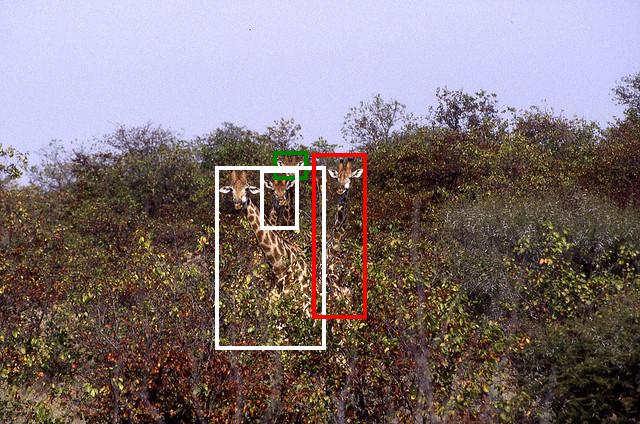}}
&
{\fontsize{9}{10}\selectfont
Is it in \textbf{left}? \hfill{No} \newline
Is it in front? \hfill{Yes} \newline
Is it in right? \hfill{No} \newline
Is it in middle? \hfill{Yes} \newline
Is it person? \hfill{No} \newline
Is it giraffe? \hfill{Yes} \newline
Is in middle? \hfill{Yes} \newline
Is in middle? \hfill{Yes} \newline
Status: \hfill{\textcolor{red}{Failure}}
}	
&
{\fontsize{9}{10}\selectfont
Is it a giraffe? \hfill{Yes} \newline
In front of photo? \hfill{Yes} \newline
In the \textbf{left half}? \hfill{Yes} \newline
Is it in the middle of photo? \hfill{Yes} \newline
Is it \textbf{to the left} of photo? \hfill{Yes} \newline
Is it to the right photo? \hfill{No} \newline
\textbf{In the left} in photo? \hfill{No} \newline
In the middle of photo? \hfill{Yes} \newline
Status: \hfill{\textcolor{green}{Success}}
}
\\ \bottomrule
\end{tabular}
\caption{Some samples generated by our improved models using REINFORCE  (left column: ``Policy Gradient'')  and Dynamic-TPG (right column: ``Tempered Policy Gradient''). The green bounding boxes highlight the target objects; the red boxes highlight the wrong guesses. }
%Note that \textsc{the Dynamic-TPG  clearly} improves the utterance qualities, compared to REINFORCE.
%\textsc{Comment: it is not obvious to me why the right columns are successes. In which sense. Can you explain?}}{It is explained the subsection "TPG Dialogue Samples+} 
\label{tab:samples}
\end{table*}

\textbf{Ablation Study:}
From Tab.~\ref{tab:results} (\#\,2 \& 3), we see that our reimplementation using Torch \cite{torch} achieves a comparable performance compared to the original TensorFlow implementation \cite{guesswhat_github}.
We use our reimplementation as the baseline.

Upon the baseline, the new QGen model with Seq2Seq structure improves the performance by about 1\%, see Tab.~\ref{tab:results} (\#\,3 \& 4).
With the Seq2Seq structure, our QGen model is trained in \textit{question level}. This means that the model first learns to query meaningfully, step by step. Eventually, it learns to conduct a meaningful dialog. Compared to directly learning to manage a strategic conversation, this bottom-up training procedure helps the model absorb knowledge, because it breaks large tasks down into smaller, more manageable pieces. This makes the learning for QGen much easier.

The next improvement is because of our new Guesser model, which uses Memory Network with two-hops attention \cite{sukhbaatar2015end}. The memory and attention mechanisms bring an improvement of 4.85\%, as shown in Tab.~\ref{tab:results} (\#\,4 \& 5). Furthermore, we train the new Guesser model additionally via REINFORCE (\#\,6). In this way, the Guesser and the QGen learn to cooperate with each other and improve the performance by another 1.34\%, as shown in Tab.~\ref{tab:results} (\#\,5 \& 6).

%In the remainder of the section, we use our models, boosted by our new QGen and Guesser trained jointly with REINFORCE as a strong baseline (\#\,6) and analyse the performance improvements achieved by the TPGs.

%\textbf{TPG Improvement:}
%\begin{figure}%[thpb], figure* for take two columns
%	\centering
%	\includegraphics[width=1.3 in]{ValidationAccuracyComparsionDuringReinforcementLearning}
%%	\missingfigure[figwidth=8cm]{Place holder}
%	\caption{Validation accuracies in reinforcement learning}
%	\label{fig:validation_curve}
%\end{figure}
Here, we take a closer look at the improvement brought by TPGs.
From Tab.~\ref{tab:results}, we see that compared to the REINFORCE-trained models (\#\,6), Single-TPG (\#\,7) with $\tau_{global} = 1.5$ achieves a comparable performance.
With two different temperatures $\tau_1 =1.0$ and $\tau_2 = 1.5$, Parallel-TPG (\#\,8) achieves an improvement of
approximately 4\%.
%and its learning curve is more stable, shown in Fig.~\ref{fig:validation_curve}.
Parallel-TPG requires more computational resources.
%\textsc{(Comment: always report improvement wrt baseline)}{OK}
Compared to Parallel-TPG, Dynamic-TPG only uses the same computational power as REINFORCE does and still gives a larger improvement by using a dynamic temperature, $\tau_t^h \in [0.5, 1.5]$. After comparison, we can see that the best model is Dynamic-TPG (\#\,9), which gives a 4.65\% improvement upon new models (\#\,6). 
%Here, we have shown that our proposed methods contribute orthogonally, in the sense that they further improve the models already boosted with advanced modules such as memory network.
%\textsc{(Comment: the term incremental is mostly used negatively as, asmost insignificant. Maybe you can rephrase)}{use orthogonally instead of incremental}
%The training curves of the validation accuracies are shown in Fig.~\ref{fig:validation_curve}, which indicates that overall TPGs converge faster and give better results.

\textbf{TPG Dialogue Samples:}
The generated dialogue samples in Tab.~\ref{tab:samples} can give some interesting insights. %in explaining why TPG methods give a better result. 
First of all, the sentences generated from TPG-trained models are on average longer and use slightly more complex structures, such as ``Is it in the middle of photo?" instead of a simple form ``Is it in middle?".
%\textsc{photo?}" \textsc{(comment: why a space?)} {it shows that ? is a token same as any other words. I deleted the space.}
Secondly, TPGs enable the models to explore better and comprehend more words. %to explore more words and comprehend them. 
For example, in the first task (upper half of Tab.~\ref{tab:samples}), both models ask about the category. The REINFORCE-trained model can only ask with the single word ``car" to query about the vehicle category. In contrast, the TPG-trained model can first ask a \emph{more general} category with the word ``vehicle" and follows up querying with a \emph{more specific} category ``trucks". These two words ``vehicle" and ``trucks" give much more information than the single word ``car", and help the Guesser model identify the truck among many cars. 
Lastly, similar to the category case, the models trained with TPG can first ask a \emph{larger} spatial range of the object and follow up with a \emph{smaller} range.
In the second task (lower half of Tab.~\ref{tab:samples}), we see that the TPG-trained model first asks ``In the left half?", which refers to all the three giraffes in the left half, and the answer is ``Yes''.  
Then it asks ``Is it to the left of photo?", which refers to the second left giraffe, and the answer is ``Yes''.
Eventually the QGen asks ``In the left in photo?'', which refers to the most left giraffe, and the answer is ``No''. 
These specific questions about locations are not learned using REINFORCE.
The REINFORCE-trained model can only ask a similar question with the word ``left". In this task, there are many giraffes in the left part of the image. The top-down spatial questions help the Guesser model find the correct giraffe. To summarize, the TPG-trained models use longer and more informative sentences than the REINFORCE-trained models.

\section{Conclusion}

Our paper makes two contributions.
Firstly, by extending existing models with Seq2Seq and Memory Networks we could improve the performance of a goal-oriented dialogue system by 7\%.
Secondly, we introduced TPG, a novel class of temperature-based policy gradient approaches.
TPGs boosted the performance of the goal-oriented dialogue systems by another 4.7\%. Among the three TPGs, Dynamic-TPG gave the best performance, which helped the agent comprehend more words, and produce more meaningful questions.
TPG is a generic strategy to encourage word exploration on top of policy gradients and can be applied to any dialog agents.

%\clearpage

\bibliographystyle{IEEEbib}
\bibliography{reference_tpg}

\end{document}